# BAYESIAN NETWORKS APPLIED TO THERAPY MONITORING

**Carlo Berzuini[1], Riccardo Bellazzi**
Dipartimento di Informatica e Sistemistica
Universita'di Pavia
27100 Pavia (Italy)

**David Spiegelhalter**
Medical Research Council
Biostatistics Unit
Cambridge CB2 2BW, U.K.

## Abstract

We propose a general Bayesian network model for application in a wide class of problems of therapy monitoring. We discuss the use of stochastic simulation as a computational approach to inference on the proposed class of models. As an illustration we present an application to the monitoring of cytotoxic chemotherapy in breast cancer.

## 1. INTRODUCTION

Of interest here is the general problem of *monitoring* and *controlling* a biomedical process over time. The basic premise is that a *model* of the process of interest is available, which allows *learning* from past data and *predicting* the future evolution of the process. Such a model may provide an intelligent system with a basis for :

(a) activating alarms which indicate significant deviations from expected progress ;
(b) using the accumulating data about the process to "learn" about the model parameters, for a more sensitive monitoring and for more accurate patient-specific predictions;

In this paper we propose a model structure that appears to be generic to the monitoring problem, and therefore likely to be useful in a wide class of clinical monitoring problems. Typical applications may involve short-term drug delivery, or medium term treatment, or long-term monitoring of chronic disease. In drug delivery the process of interest is the temporal variation of the drug concentration in body compartments, and the goal is to suggest dosage adjustments that are necessary to achieve concentrations which lie between specific desired limits.

In the management of a chronic disease the monitored process is typically the patient's progression through stages of the disease, and the goal of the monitoring is to predict possible risks and benefits of changes in therapy. An application to medium-term treatment monitoring is used in this paper as an illustration.Clinical practice in medium- or long-term monitoring situations often involves simple rules in which an observed change in the process provokes a corresponding action. Usually these rules are grossly inappropriate for particular patients. Our approach should allow a progress with respect to such a scheme, in that the rules would be replaced by an adaptive model, or the model might be used to evaluate the reliability of the rules.

In the field of drug dosage individualization a Bayesian approach has previously been suggested (eg. D'Argenio 1988, Sheiner 1982). However a flexible framework for model development and for inference computations in this area has not previously been proposed.

In this paper we use a *Bayesian network* both as a representation of the model and as a computational framework for inference. Bayesian networks are able to crystallize in a graphical representation the rich mixture of causal knowledge and conditional independence assumptions that underlies a complex probabilistic model. At a high level, the network will display the way in which the basic sub-systems relate and interlock to form the complete mosaic that represents the process of interest. This is particularly useful in models that involve development over time. At a lower level, the network will display in detail the conditional independence relationships among individual variables in each portion of the global model.

The computation of inferences on the model can be conveniently performed using a stochastic simulation algorithm, called the Gibbs sampler [Geman & Geman, 1984] [Pearl, 1987] [Gelfand, 1988] [Henrion, 1990] [Shachter, 1990]. The essence of this algorithm consists of sampling the joint posterior distribution of unobserved

[1] *Mailing address:* Carlo Berzuini, Dipartimento di Informatica e Sistemistica, via Abbiategrasso 209. 27100 Pavia (Italy). FAX: (39) 382-422881

variables/parameters in the model. The result is an *approximation* of the desired posteriors.

In alternative, one might use algorithm for the exact calculation of posteriors (see, for example, Lauritzen and Spiegelhalter, 1988). However, algorithms for "exact" probability propagation may raise difficulties in certain situations. The first is when the graph is too tightly connected, which will particularly occur when population parameters are not precisely specified. The second is when continuous variables are involved: "exact" propagation currently requires that these variables are discretized, and a recent proposed "exact" method for propagating probabilities on networks with mixed qualitative and quantitative variables [Lauritzen, 1990] seems to assume heavy restrictions on the distributions involved. A compile-time discretization may be hampered by the difficulty of deciding the most appropriate binning and bounding; a run-time ("dynamic") discretization may led to a significant overhead, particularly if the involved distributions are complex.

Thanks to the availability of standard routines for sampling from continuous distributions [Ripley, 1987], stochastic simulation takes advantage of what we know to hamper the applicability of exact methods, namely the continuous nature of the variables, particularly non-Gaussian distributions.

abstraction, a node at a given level representing one or more nodes at lower levels. This greatly facilitates insight into the model.

In this section we consider a top-level view of our model, which is shown in Fig.1.

We suppose that a response relationship of interest has been observed on $N$ cases, typically patients, that are somehow similar. The $N$-th patient, called the *target case*, is currently under observation. Node $X$ represents future evolution of the target case, conditional on a plan for future therapeutic action.

Usually there will be a substantial patient *heterogeneity* in their response to the therapy, and part of this variation will be "explained" by a set $z$ of covariates (eg. age, sex) that affect an individual's response characteristics. We let $z_i$ denote the covariate vector for the $i$-th patient. To handle patient heterogeneity, we introduce *patient-specific* unknown *response parameter vectors*, $\theta_1,...,\theta_N$. We regard the generic $\theta_i$ as drawn from a density $p(\theta_i | z_i, \beta, \kappa)$, where the unknown parameters $\beta$ represent effects associated with the individual covariates, and $\kappa$ is a vector of distributional parameters. For notational economy we pool $\beta$ and $\kappa$, which we call the *hyperparameters*, into a single parameter vector $\pi = (\beta, \kappa)$,

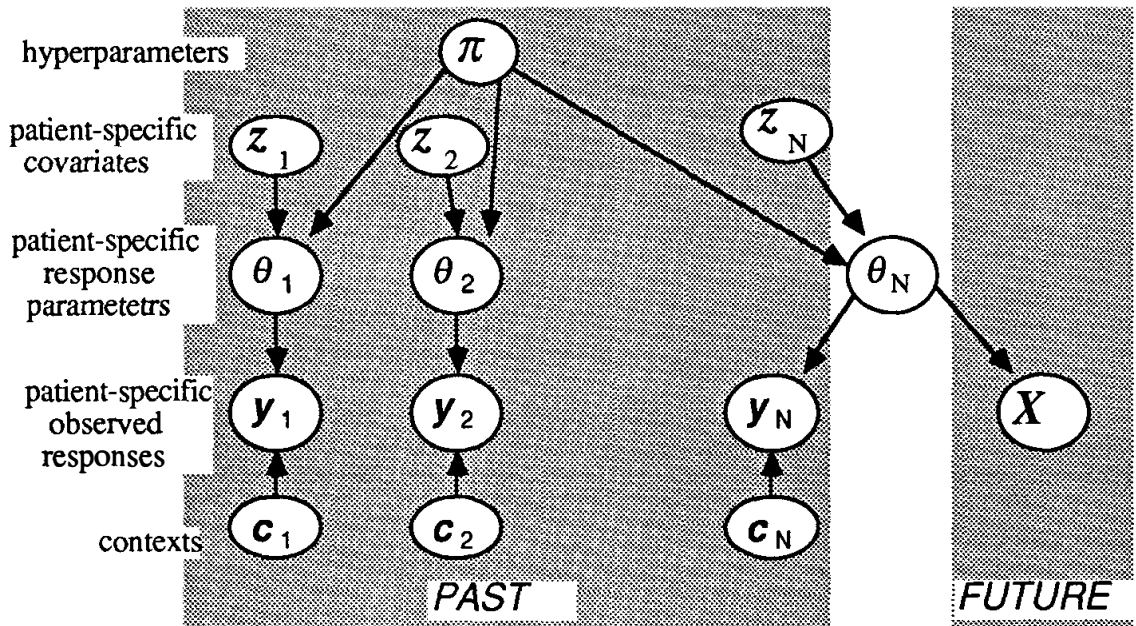

Figure 1. High-level view of the general Bayesian network model for therapeutic monitoring

## 2. HIGH-LEVEL VIEW OF THE MODEL

One of the advantages of Bayesian networks is that they can be developed at progressively higher levels of detail. One then obtains network models at different levels of called the *hyperparameter vector*. We assume that the generic $\theta_i$ completely defines the response characteristics for the $i$-th patient.

Therefore, the vector of the actual observations on the response of the $i$-th patient, denoted as $y_i$, can be viewed as a random vector drawn from the density $p(y_i$

|$\theta_i, c_i$), where the vector $c_i$ summarizes the *known* values of a set of context variables, describing *contingent* conditions such as dose schedules, observation times, or contingent aspects of the patient such as, for example, values of physiological quantities at drug administration.

## 3. INFERENCE

The inference procedure starts with using "population" information:

$$D = \{y_1,...,y_{N-1}, c_1,...,c_{N-1}, z_1,..., z_{N-1}, z_N\}$$

to shape a prior distribution for the unknown parameters $\theta_N$ for the target case. Then, case-specific response data $\{y_N, c_N\}$ are used to update such prior into a posterior distribution that reflects all the current information about the target case, while "borrowing strength" from the whole past experience $D$. A classical maximum-likelihood approach tends to yield a poor estimate of $\theta_N$ or of linear transformations of it if the record $\{y_N, c_N\}$ contains poor information. By contrast, in our approach the estimates of those components or transformations of $\theta_N$ which are poorly estimated from the patient's-specific data will be drawn towards the population average, so that more reasonable values should be obtained.

A further benefit of our approach is that the estimation of $\theta_N$ is done through a *sequential updating* procedure: as new data from the monitored patient (or from data-base patients) become available they are incorporated to yield a revised posterior distribution of $\theta_N$, and thus a revised predictive distribution of $X$. Significant changes in the patient's behaviour, possibly pointing to important patho-physiological events, should be mirrored by concomitant patterns of change of the posterior of $\theta_N$.

The inference involves four basic steps of probability propagation over the net of Fig.1. These steps are separately described in the following. Below the description of each single step, the relevant portion of the net of Fig. 1 is shown.

- **population updating**: initially we pretend to be completely uninformed about the value of the hyperparameters $\pi$, and represent this "ignorance" by associating with node $\pi$ a "vague" *hyperprior* distribution, denoted $p_0(\pi)$. Then information $D = \{y_1,...,y_{N-1}, z_1,..., z_{N-1}, c_1,..., c_{N-1}\}$ is used to update $p_0(\pi)$, using probability propagation over the network below, into an *a posteriori* distribution $p_1(\pi | D)$, called the *population distribution* of $\pi$;

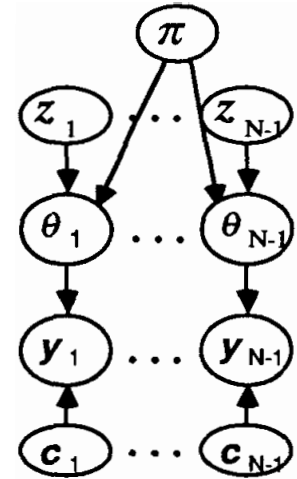

- **collapsing**: prior to considering the specific data $\{y_N, c_N\}$, all we know about the target case is the corresponding covariate vector $z_N$ and the fact that he/she/it is a member of the same population from which the remaining $N-1$ data base cases were drawn. This leads to regard the parameters $\theta_N$ for the target case as those of a generic population individual with covariables $z_N$, and to view $\theta_N$ as drawn from the distribution:

$$p_0(\theta_N | z_N, D) = \int_\Omega p_\theta(\theta_N | \pi, z_N) p_1(\pi | D) d\pi \quad (1)$$

where $\Omega$ denotes relevant domain of $\pi$.

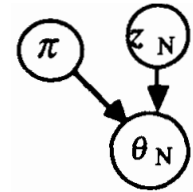

- **case-specific updating**: past data $\{y_N, c_N\}$ obtained from the target case are used to update the prior $p_0(\theta_N | z_N, D)$ into a posterior distribution $p_1(\theta_N | z_N, y_N, c_N, D)$ which, by Bayes' theorem, is given by:

$$p_1(\theta_N | z_N, y_N, c_N, D) \propto p_0(\theta_N | z_N, D) \times p(y_N | \theta_N, c_N) \quad (2)$$

and that describes our uncertainty about the response parameters for the target case, *after considering all relevant (population- and patient-specific) available information*.

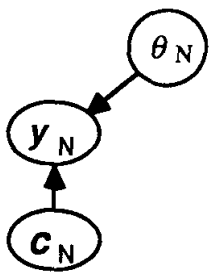

**prediction**: $p_1(\theta_N | z_N, y_N, c_N, D)$ is used to obtain a predictive distribution of $X$, denoted $p_P(X)$. Future patient's evolution is predicted under different hypothesized decision plans for a more rational decision. $p_P(X)$ is defined by :

$$p_P(X) = \int(\Sigma; p(X | \theta_N) p_1(\theta_N | z_N, y_N, c_N, D) \, d\theta_N \quad (3))$$

where $\Sigma$ denotes relevant domain of $\theta_N$.

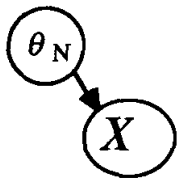

## 4. COMPUTING THE INFERENCES VIA STOCHASTIC SIMULATION

The computations required in the above described inference phases can be conveniently performed by *sampling* techniques. For example, collapsing requires the integration (1). This can be performed by first drawing a sample $\pi^{(0)}$ from $p_1(\pi | D)$, then a sample $\theta^{(0)}$ from $p_\theta(\theta_N | \pi^{(0)}, z_N)$, then a new sample $\pi^{(1)}$ from $p_1(\pi | D)$ and a new sample $\theta^{(1)}$ from $p_\theta(\theta_N | \pi^{(1)}, z_N)$, and so on. This resampling is repeated a high number of times, say $L$. At the end, one can straightforwardly use the set of generated samples $\{\theta^{(0)}, \theta^{(1)}, ..., \theta^{(L)}\}$ to calculate any marginal or summary of $p_0(\theta_N | z_N, D)$, eg. selected percentiles of the distribution of a given component of $\theta_N$.

Note that the densities that appear in (1)-(2) are multivariate, which seems to imply that the sampling steps involved in collapsing and case-specific updating must be actually performed on multivariate densities. As a matter of fact, one may draw samples from a multivariate distribution by actually sampling univariate distributions. A first possibility is to exploit conditional independence. For example, if the components of $\pi$ are conditionally independent given $D$, then one can draw samples from the multivariate density $p_1(\pi | D)$ by actually sampling individual densities $p_1(\pi_h | D)$.

Case-specific updating and prediction are conveniently carried out by an iterative stochastic simulation algorithm, called the *Gibbs sampler*, that uses a Bayesian network representation of the relationships among variables $(\theta_N, z_N, y_N, c_N)$ as the computational framework for the necessary calculations. Such a network will obviously depend on the specific application, since it describes in all details the structure of the conditional independencies among individual variables in the problem.

Perhaps the key reference to Gibbs sampling is Geman & Geman [Geman & Geman, 1984], who discuss its application to image analysis. A thorough review of the method is given by Gelfand & Smith [Gelfand, 1988]. Within the AI literature, the method has been explored, under the name of *Markov stochastic simulation*, by several authors, eg. Pearl [Pearl, 1987] and Henrion [Henrion, 1990].

To perform the Gibbs sampling, initial values are assigned to each unobserved variable in the network. Then, for each unobserved variable in turn, the current value is replaced by a value drawn from the full conditional distribution of that variable given the current values of its "neighbours" in the net (parents, children and parents of the children, observed and unobserved). This resampling is repeated many times. Under certain regularity conditions, Geman & Geman [Geman & Geman, 1984] show that the resampling process is an irreducible Markov chain that converges to an equilibrium distribution given by the full posterior distribution of the unknown variables. Thus, after the sampling, any posterior summaries or marginal components of such multivariate posterior can be straightforwardly calculated from the set of generated samples.

## 5. SPECIFIC MODEL FOR CYTOTOXIC CHEMOTHERAPY MONITORING IN BREAST CANCER

As an illustration example, we consider an application to the monitoring of patients affected by breast-cancer who are being given cycles of post-operative cytotoxic chemotherapy. From a clinical perspective, it is desirable that the patient receives an adequate dose, while guarding excessive toxicity which leaves the patient vulnerable to bouts of infection.

We need a low-level network model that describes the relationships among individual response parameters, context variables and response observations in this specific application. The idea is one of introducing suitable and reasonable conditional independence assumptions among the variables in order that the problem can be decomposed into manageable sub-problems, and that subsequent quantification of the model is straightforward.

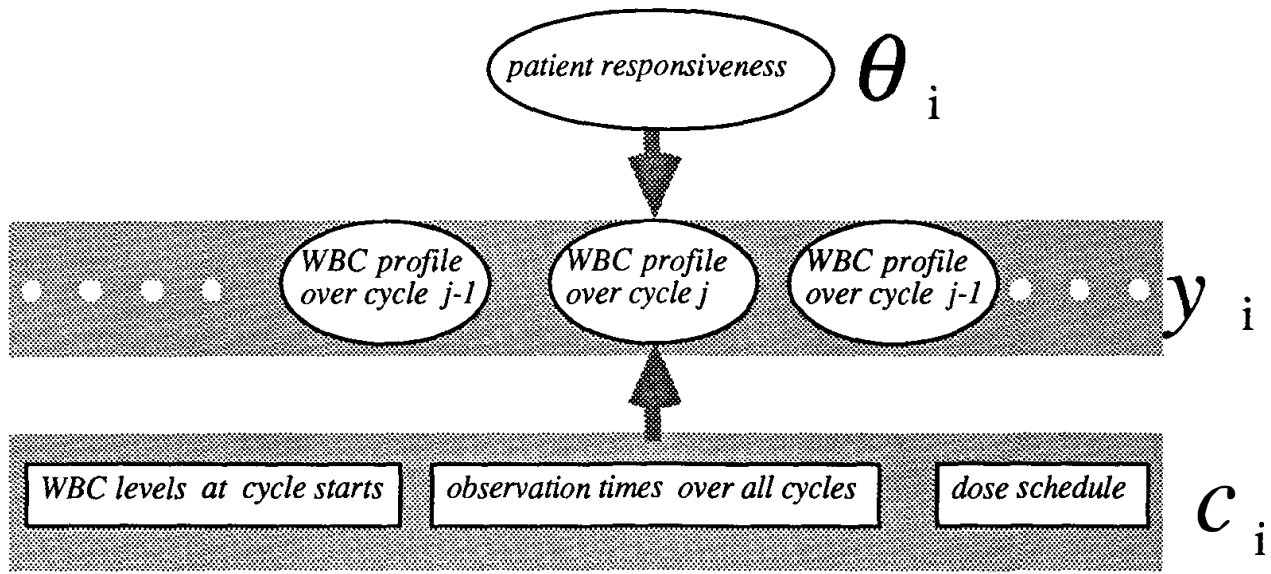

Figure 2. Intermediate-level view of a Bayesian network for the monitoring of cytotoxic chemotherapy in breast cancer. The figure shows only a portion of model for the generic $i$-th patient. A big shaded arrow pointing from a layer to another means that each node in the first layer sends an arrow to each node in the latter layer.

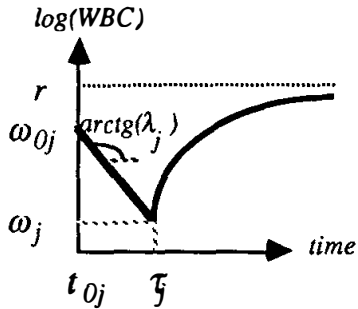

Figure 3. log-WBC-count profile over j-th cycle for a generic patient

Our proposed model, which has been described in more detail by Bellazzi et al (1991), is shown in Fig.2. This model represents the assumption that observations on individual cycles of treatment are conditionally independent given some patient-specific parameters, such as those representing the sensitivity and responsiveness of the patient.

We assume that the level of white blood cell (WBC) count is the most important aspect of bone-marrow toxicity. High toxicity is reflected by a low WBC count. Several features might be straightforwardly combined with the WBC count in our model, but nothing essential is lost if we restrict to a simple model. We let $w_{kj}$ denote $k$-th log-WBC-count measurement within the $j$-th cycle of treatment, and let $t_{kj}$ denote the time of this measurement, taken from the instant at which the drug is administered at the beginning of the cycle. The stylized profile shown in Fig. 3 is assumed to be a realistic model of the log-WBC-count profile over the generic treatment cycle.

The intercept $w_{0j}$ represents the log-WBC count at drug administration. The downward slope is denoted as $\lambda_j$, the recovery rate as $\gamma$ (assumed independent of the cycle) and the change point as $\tau_j$. The minimum log-WBC level reached during the cycle $j$, denoted as $\omega_j$, is a simple function of $\{w_{0j}, \lambda_j, \tau_j\}$.

By adding gaussian white noise errors $e_{kj}$, $k=1,...,5$; $j=1,...,NC$, we obtain the following longitudinal regression model for the log-WBC profile over treatment cycle $j$ :

$$\lambda_j = k \times dose_j \times \alpha$$
$$\omega_j = w_{0j} - k \times dose_j \times \alpha \times \tau \quad (4)$$
$$w_{kj} = w_{0j} - \lambda_j \times (t_{kj} - t_{0j}) + e_{kj} \quad \text{if} \quad 0 < t_{kj} < \tau$$
$$w_{kj} = \omega_j + (r - \omega_j)(1 - exp\{-\gamma \times (t_{kj} - \tau)\}) + e_{kj} \quad \text{if} \quad \tau < t_{kj}$$
$$e_{kj} \sim No(0, \sigma^2)$$

with :

| | |
|---|---|
| $k$ | observation within cycle ($k=1,...,5$) |
| $j$ | cycle number |
| $e_{kj}$ | realisation of a gaussian random variable with mean $0$ and variance $\sigma^2$; |
| $dose_j$ | standardized dose = actual dose given to the patient at cycle $j$ divided by square meters of body surface area ; |

40   Berzuini, Bellazzi, and Spiegelhalter

α     patient's sensitivity to the drug ($\alpha$ =1 : normal patient; $\alpha$ =1.5 : sensitive patient; $\alpha$ =2 : very sensitive patient );

k     expected fall in the log-WBC count count in unit time, for a normal patient given a unit dose of treatment ;

r     "normal" log-WBC count (the value at the start of the therapy).

$$\pi_{.1} + \pi_{.2} + \pi_{.3} = 1.$$

The parameters $\pi_{.m}$, $m$ =1,..., 3 are hyperparameters. The parameter $\sigma$ is more conveniently treated as continuous. A suitable class of prior distributions for $\sigma$ is the inverse gamma :

$$1/\sigma \sim \Gamma(a, b) \qquad (7)$$

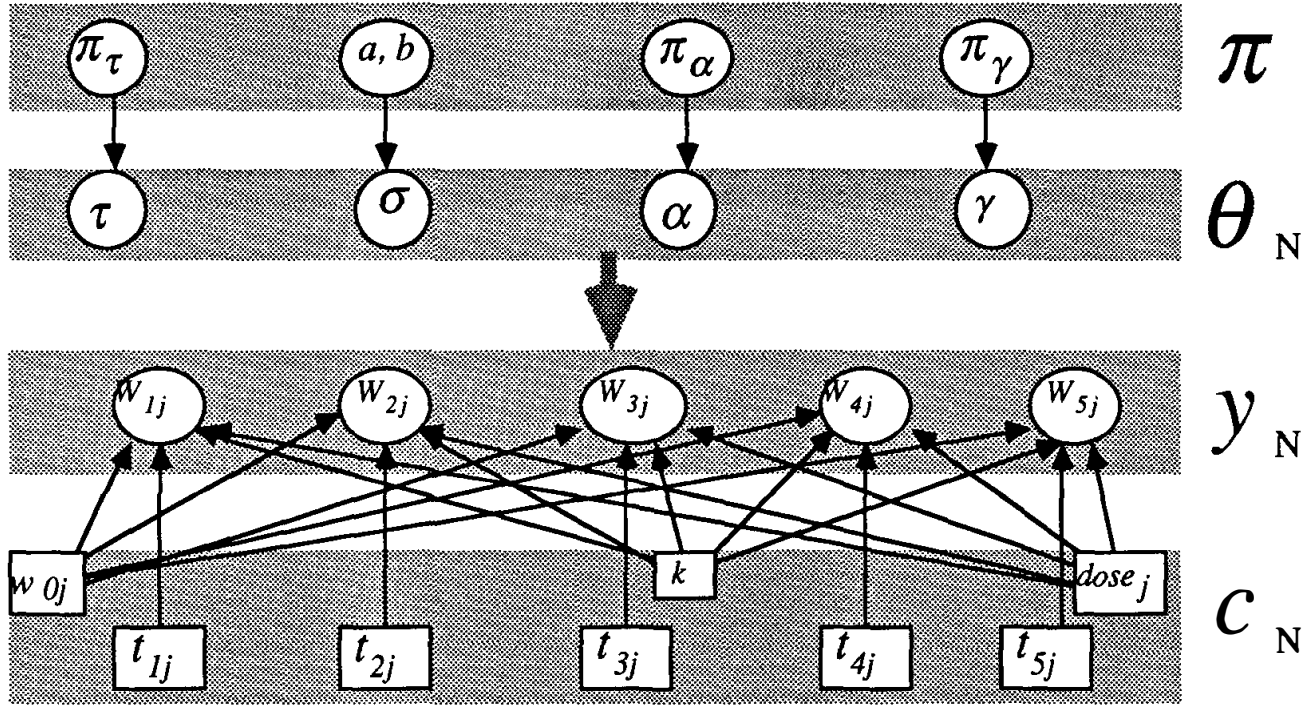

Figure 4 Bayesian Network representation of our patient-model for toxicity. Layers (from top to bottom) contain : hyperparameters, response parameters, response observations (WBC counts), context variables. A big shaded arrow pointing from a layer to another means that each node in the first layer sends an arrow to each node in the latter layer.

From (4) it follows that the conditional distribution attached to the generic observed log-WBC-count $w_{kj}$ is gaussian, and specified by :

$p(w_{kj}|$ all other variables ) =

$=No(w_{0j} - k\ dose_j\ \alpha(t_{kj}-t_{0j}), \sigma^2)$     if $\tau_j < t_{kj}$ (5)

$=No(\omega_j+(r\ -\omega_j)(1-exp\{-\gamma\times(t_{kj}-\tau)\}), \sigma^2)$ if $\tau_j \geq t_{kj}$

A complete Bayesian formulation of the model requires us to specify conditional distributions of response parameters { $\alpha, \gamma, \tau, \sigma$ } given a suitable set of hyperparameters. We decided to take $\alpha, \gamma, \tau$ as 3-level discrete random variables with respective prior probability mass functions:

$p(\alpha =m )=\pi_{\alpha m}$, $p(\gamma =m )=\pi_{\gamma m}$, $p(\tau =m )=\pi_{\tau m}$     (6)

$m$ =1,..., 3

which preserves conjugacy with respect to (5). The quantities $a, b$ are hyperparameters.

Finally we need to attach prior distributions to the hyperparameters. A suitable prior for each hyperparameter is the *uniform* distribution over a suitable interval. Such a "vague" hyperprior expresses our *a priori* total ignorance about such hyperparameters.

The qualitative relationships implicit in equations (4)-(7) are depicted in the Bayesian Network shown in Fig. 3. The meaning of a generic arrow $x \rightarrow y$ may be interpreted to be "$x$ *contributes to define the distribution from which $y$ is generated*". The distributional specifications described above provide a complete quantification of the network, in that they specify the conditional distribution of each random variable given its parents in the network.

Note that the network contains a node for each single observation on WBC. Implicit in the network structure is the assumption that these observations are conditionally independent given the response parameters and the context variables. This is correct as long as one neglects correlation among errors $e_{kj}$.

One could acknowledge the fact that response parameters may vary from cycle to cycle by introducing cycle specific parameters $\alpha_j$, $\sigma_j$ and so on, and impose a structure (eg. a time series model) that expresses the temporal relatedness among these, but we shall not pursue this here.

### 5.3 Numerical results

A computer program, called GAMEES (Bellazzi, 1991), has been developed for general purpose Bayesian network modelling and inference using stochastic simulation. The network structure is entered into the computer through a graphical interface. The program requests the user to enter only essential aspects of the overall multi-level network structure. The program then internally generates all the implied repetitive structure. GAMEES incorporates a wide library of sampling routines and sampling control strategies for carrying out all the inference steps described in section 3 and processing the generated samples. The population-updating phase is carried out through a connection between the Bayesian network and a patient data-base.

By means of GAMEES, we applied the model of Fig.3 to the treatment records of 11 patients undergoing breast-cancer post-operative chemotherapy. These records contained data from 2 to 3 treatment cycles per patient, each cycle yielding at most 5 data points.

*Population updating* and *collapsing* yielded population priors for response parameters. In particular, for parameter $\alpha$ we obtained: $p_1(\alpha =1)= 0.33$ and $p_1(\alpha=1.5)= 0.36$ and $p_1(\alpha=2)= 0.31$, indicating that patients with "normal" sensitivity are slightly more frequent than others.

The data from the chosen *target patient* spanned 5 treatment cycles. Fig.5a shows the data from the target patient superimposed to a "cloud of points" that represents a collection of profiles simulated using population distributions for the response parameters. The profiles appear to be very dispersed, reflecting high heterogeneity within the population. It is apparent that the target patient is much more sensitive than the population average, and that a prediction of the response for the target patient based on population distributions of the unknown model parameters would be inappropriate.

*Patient-specific updating* was first carried out on the basis of first cycle data from the target patient. 500 Gibbs sampling runs were employed. The convergence of the algorithm was monitored by plotting profiles of values generated sequentially during the Gibbs sampling, one plot for each of the four unknown response parameters. A reasonably stationary behaviour of the four profiles was achieved after 100 runs, although substantial cyclicity persisted. Subsequent samples were picked up at a rate of one out of every five to reduce autocorrelation. The obtained predictive distribution of subsequent WBC count evolution over cycles 2 to 5 is represented in Fig.5b by a cloud of points. Big points represent actual observations. One may note that the prediction fits the actually observed patient's data much better than the population-based prediction of Fig. 5a. Moreover, the good fit between actual observations made in cycles 2 to 5 and the corresponding prediction indicates that the patient is "responding as expected". Any discrepancy may lead to a decision (e.g. to change therapy) and can also be used to update response parameters.

Fig.5c displays the prediction of the patient's response over cycles 4 and 5, based on data of cycles 1 to 3.

Fig.5d shows the evolution of the estimated posterior distribution for parameter $\alpha$ over the cycles of therapy. Note that the sequential updates gradually shift our belief towards the highest value for parameter $\alpha$, thus strengthening our opinion that the patient is "very sensitive". Another useful summary would be a plot of percentile curves for posteriors of interest.

## 7. CONCLUSIONS

We have described an approach to therapy monitoring that we hope of potential usefulness in many areas of clinical monitoring, such as short-term drug delivery, medium-term therapy managing and long-term monitoring of chronic diseases. An important characteristic of this approach is the capability of learning at a population level and adapting to a specific patient.

In the future we intend to explore, in different application areas, the possibility of developing these models into even larger networks by extending them to include processes that develop on different temporal scales, such as when short-term drug delivery and medium-term clinical outcomes are taken into account jointly. A suitable conditional independence structure would make a large network model manageable, allowing its various portions to be studied individually, so that different areas of expertise might be cooperatively brought to bear.

An investigation of the possibilities to combine the modelling approach proposed in this paper with other approaches, such as compartmental modelling, provides much scope for further work. In particular, compartmental modeling of the underlying drug metabolism might provide an essential tool for achieving meaningful and parsimonious parametrisations of the response relationship to be modelled. This would allow pathophysiological knowledge to be brought to bear, and would be particularly useful when multiple response curves, generated by the same underlying metabolic dynamics are involved.

## Acknowledgements

M.Leaning made essential contributions to the development of the application. R.Bianchi and S.Quaglini implemented (together with R.Bellazzi) the GAMEES program. This work was partly supported by an EEC grant (AIM Project 1005).

## References

Bellazzi, R., Berzuini, C., Quaglini, S., Spiegelhalter, D., Leaning, M. (1991), Cytotoxic chemotherapy monitoring using stochastic simulation on graphical models. *to appear on the Proceedings of the AIME-91 Conference, Maastricht, Nederland, july 1991*.

D'Argenio, D.Z. and Maneval, D.C. (1988). Estimation approaches for modeling sparse data systems, *Proceedings of the IFAC Symposium on Modeling and Control in Biomedical Systems, 6-8 april, Venice, Italy* (C.Cobelli and L.Mariani, eds.). Pergamon Press, New York.

Gelfand, A. and Smith, A.F.M. (1988), Sampling based approaches to calculating marginal densities. *Technical Report*, University of Nottingham, U.K.

Geman, S. & Geman, D. (1984), Stochastic relaxation, Gibbs distributions, and the Bayesian restoration of images. *IEEE Transactions on Pattern Analysis and Machine Intelligence*, PAMI-6, pp. 721-741.

Henrion, M. (1990), An Introduction to Algorithms for Inference in Belief Nets, in: *Uncertainty in Artificial Intelligence 5* (M.Henrion, R.D.Shachter, L.N. Kanal and J.F.Lemmer eds.), Elsevier Science Publishers B.V. (North-Holland).

Lauritzen, S.L. & Spiegelhalter, D.J. (1988) Local computations with probabilities on graphical structures and their application to expert systems (with discussion). *J. Roy. Statist. Soc., B*, 50, 157-224.

Lauritzen, S. (1990), Propagation of probabilities, Means and Variances in Mixed Graphical Association Models, *Technical Report R 90-18*, Institute for Electronic Systems, Department of Mathematics and Computer Science, University of Aalborg, Denmark, april 1990.

Pearl, J. (1987), Evidential reasoning using stochastic simulation of causal models. *Artificial Intelligence*, 32, no.2, 245-252.

Ripley, B.D. (1987), *Stochastic Simulation*, Wiley, New York.

Shachter, R.D., Peot, M. (1990), Simulation Approaches to General Probabilistic Inference on Belief Networks, in: *Uncertainty in Artificial Intelligence 5* (M.Henrion, R.D.Shachter, L.N. Kanal and J.F.Lemmer eds.), Elsevier Science Publishers B.V. (North-Holland).

Sheiner, L.B., Beal, S.L., (1982), Bayesian individualisation of pharmakokinetics: simple implementation and comparison with non Bayesian methods. *J. Pharm. Sci.*, 71, 1344-1348.

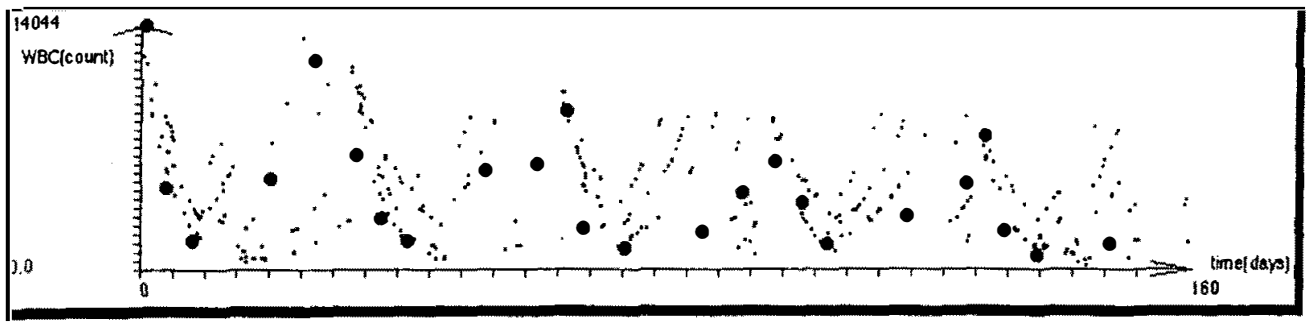

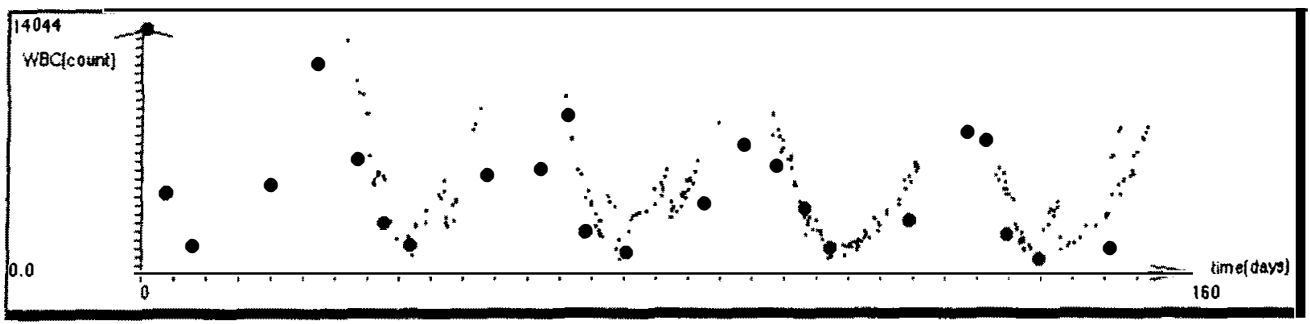

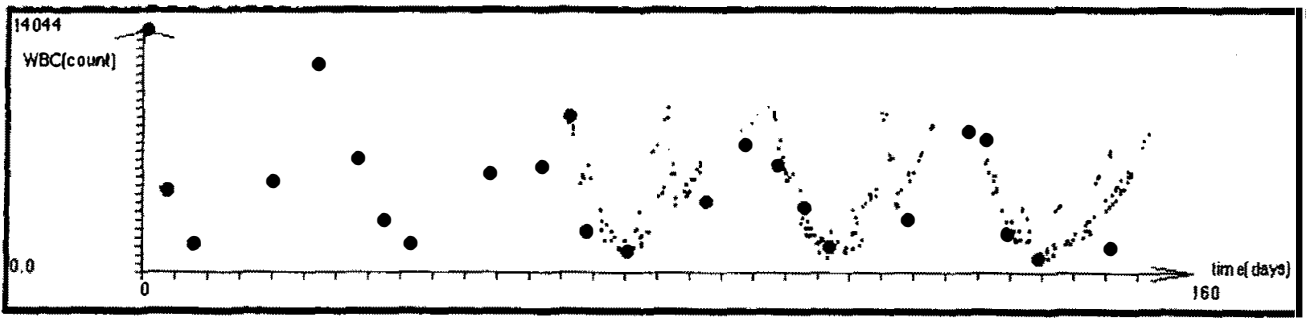

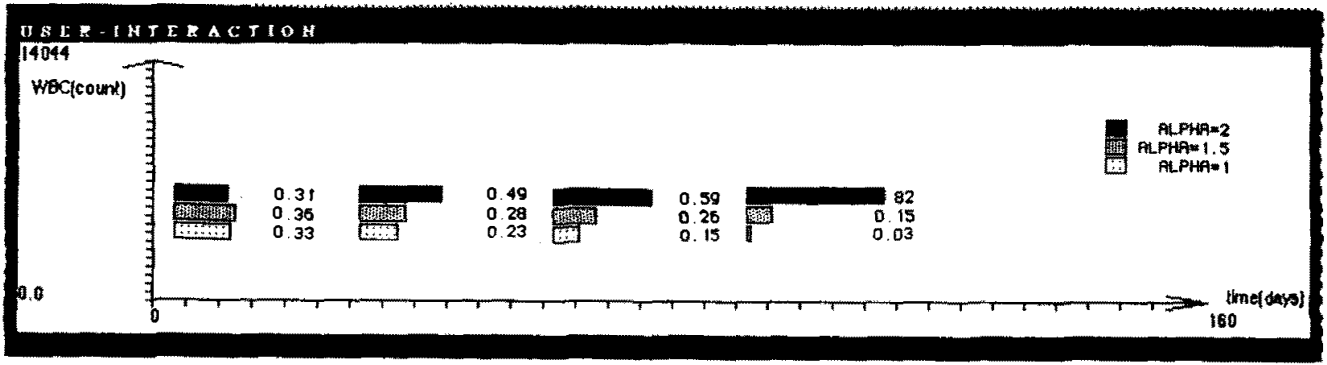